\documentclass[sigconf,nonacm]{acmart}

\setcopyright{none}
\renewcommand\footnotetextcopyrightpermission[1]{}
\usepackage{tikz}
\usetikzlibrary{arrows.meta,positioning,calc}
\usepackage{listings}

\AtBeginDocument{%
  }

\acmConference[KDD '26 Workshop on AI Agents]{KDD 2026 Workshop on AI
  Agents}{August 2026}{}
\acmYear{2026}
\copyrightyear{2026}

\makeatletter
\newcommand\boldparagraph{\def\@toclevel{4}%
  \@startsection{paragraph}{4}{\parindent}%
  {-.5\baselineskip \@plus -2\p@ \@minus -.2\p@}%
  {-3.5\p@}%
  {\ACM@NRadjust{\bfseries\@adddotafter}}}
\makeatother

\definecolor{lstbg}{rgb}{0.97,0.97,0.97}
\definecolor{lstframe}{rgb}{0.80,0.80,0.80}
\definecolor{lststr}{rgb}{0.76,0.52,0.06}   
\definecolor{lstcom}{rgb}{0.55,0.55,0.55}   
\lstdefinestyle{toolcall}{%
  language=Python,
  basicstyle=\ttfamily\footnotesize\color{black},
  backgroundcolor=\color{lstbg},
  frame=single,
  framesep=5pt,
  rulecolor=\color{lstframe},
  keywordstyle=\color{black},
  stringstyle=\color{lststr},
  commentstyle=\color{lstcom},
  identifierstyle=\color{black},
  showstringspaces=false,
  breaklines=true,
  breakatwhitespace=false,
  breakindent=0pt,
  postbreak=\mbox{\textcolor{lstcom}{$\hookrightarrow$}\space},
  columns=fullflexible,
  keepspaces=true,
  aboveskip=6pt,
  belowskip=2pt,
  xleftmargin=2pt,
  xrightmargin=2pt,
}

\begin{document}

\title{Beyond Next-Token Prediction: An RLVR Proof of Concept for
  Tool-Use Agents on Atlassian Workflows}

\author{Karthikeya Aditya Vissa}
\authornote{Equal contribution.}
\affiliation{\institution{Centific}\country{}}
\email{karthikeyaaditya.v@centific.com}

\author{Sankalp Mane}
\authornotemark[1]
\affiliation{\institution{Centific}\country{}}
\email{sankalp.mane@centific.com}

\author{Ananya Mantravadi}
\affiliation{\institution{Centific}\country{}}
\email{ananya.mantravadi@centific.com}

\author{Harshit Rajgarhia}
\affiliation{\institution{Centific}\country{}}
\email{harshit.rajgarhia@centific.com}

\author{Abhishek Mukherji}
\affiliation{\institution{Centific}\country{}}
\email{abhishek.mukherji@centific.com}

\renewcommand{\shortauthors}{Aditya Vissa, Mane, Mantravadi, Rajgarhia, and Mukherji}

\begin{abstract}
Large language models are trained to predict the next token, not to act
inside a specific API. In niche enterprise SaaS workflows --- where
success means hitting the right endpoint with the right nested arguments
in the right order --- this objective mismatch shows up as silent
failures: dropped required fields, hallucinated tools, or early stops
after a single read. We ask whether Reinforcement Learning with
Verifiable Rewards (RLVR), applied directly in the target environment,
closes the gap. As a proof of concept we build a suite of five synthetic
environments emulating the Jira REST v3 and Confluence v2 APIs at schema
fidelity; rewards are computed entirely from the tool-call trace, with no
live API, no learned judge, and no human label in the loop. Scoring
prompted Qwen3-1.7B and Qwen3.5-4B on the same checkers that drive GRPO
training, we find that on the four scenarios whose rewards are
non-degenerate the RL-trained policy lifts average reward from a
4B-baseline range of 0.35--0.92 to 0.95--1.00, with the largest single
gain on Confluence page creation ($0.35 \rightarrow 1.00$). We position
this as a preliminary step toward outcome-optimised small models for
niche enterprise APIs, and foreground two limitations a workshop reader
should weigh: hand-crafting verifiable rewards does not scale beyond the
handful of endpoints reported here, and one of our five scenarios
(ticket-transition) has a saturating reward shape that the prompted 4B
already maxes out.
\end{abstract}


\maketitle

\begin{figure*}[t]
\centering
\begin{tikzpicture}[
  >={Stealth[length=2.4mm]},
  box/.style={draw, rounded corners=2pt, align=center, inner xsep=6pt,
    inner ysep=5pt, minimum height=10mm, font=\small},
  lbl/.style={font=\scriptsize\itshape, inner sep=2pt},
  node distance=6mm,
]
\node[box, fill=gray!12] (prompt) {User prompt\\[1pt]{\scriptsize(6--12 / scenario)}};
\node[box, fill=red!8, right=of prompt] (policy) {Qwen3 policy\\[1pt]{\scriptsize(1.7B / 4B, BF16)}};
\node[box, fill=blue!8, right=of policy] (api) {Synthetic API\\[1pt]{\scriptsize(Jira v3 + Conf v2)}};
\node[box, right=of api] (trace) {Tool-call trace\\[1pt]{\scriptsize(calls + args)}};
\node[box, fill=green!12, right=of trace] (reward) {Verifiable reward\\[1pt]{\scriptsize(R1 args + R2 order + R3 penalties)}};
\node[box, fill=yellow!18, below=12mm of trace] (grpo) {GRPO update\\[1pt]{\scriptsize(group-relative)}};

\draw[->] (prompt) -- (policy);
\draw[->] (policy) -- node[lbl,below]{tool call} (api);
\draw[->] (api) -- (trace);
\draw[->] (trace) -- (reward);
\draw[->] (api.north) to[out=120,in=60] node[lbl,above]{response} (policy.north);
\draw[->] (reward.south) |- node[lbl,above,pos=0.78]{scalar reward} (grpo.east);
\draw[->,dashed] (grpo.west) -| node[lbl,above,pos=0.22]{parameter update} (policy.south);
\end{tikzpicture}
\caption{The RLVR training loop. A user prompt enters a Qwen3 policy that
emits tool calls into a synthetic, schema-faithful copy of the Jira REST
v3 / Confluence v2 APIs; the resulting tool-call trace is scored by a
hand-designed verifiable reward (per-argument correctness,
validate--mutate--verify structural bonuses, hallucination and
excess-call penalties); GRPO updates the policy on the per-prompt group
of rollouts. No live API, no learned judge, and no human label are in the
loop.}
\Description{A horizontal pipeline of five boxes: User prompt, Qwen3
policy, Synthetic API, Tool-call trace, and Verifiable reward. The reward
feeds a scalar into a GRPO update box, which sends a parameter update back
to the Qwen3 policy, closing the loop.}
\label{fig:loop}
\end{figure*}
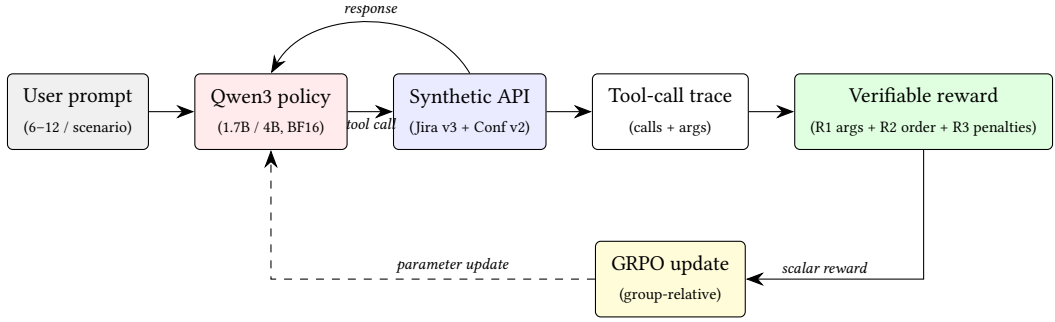

\section{Introduction}
\label{sec:intro}

LLM-driven agents are moving from chat into workflows: opening tickets,
transitioning issues, updating wikis. A growing literature treats tool
invocation as a first-class capability to be elicited via supervised
fine-tuning or RL on real APIs~\cite{qin2024toolllm,schick2023toolformer,wang2024codeact,yao2023react},
yet public RL benchmarks for agents target the open web~\cite{zhou2024webarena},
software engineering~\cite{jimenez2024swebench}, or broad agentic
suites~\cite{liu2024agentbench,trivedi2024appworld}. Niche enterprise
SaaS workflows --- narrow, schema-heavy, dominated by the ``check the
resource, then mutate it'' pattern --- get comparatively little
attention.

The deeper issue is an objective mismatch: LLMs are optimised against
next-token likelihood on internet-scale text, not to act inside a
particular API surface. Asked to create a Jira sub-task or a Confluence
page in our suite, a prompted Qwen3.5-4B knows the rough shape of the call
but fills slots wrong, omits required fields, or stops after the first
read --- behaviours that read as fluent under a token-prediction
objective and score poorly under any check that grounds the output back to
the environment (our prompted-baseline scores in
Section~\ref{sec:results} bear this out). \textit{Reinforcement Learning
with Verifiable Rewards} (RLVR)~\cite{cobbe2021gsm8k,deepseek2025r1,lambert2024tulu3,shao2024deepseekmath}
gives up reward modelling in favour of programmatic checkers wherever
correctness can be inspected. Tool-use is naturally amenable: an agent's
output is a sequence of structured tool calls whose argument values and
ordering can be inspected directly. We exploit that observation to train
agents against outcome-level rewards in synthetic copies of the target
environment, with no live API, learned judge, or human label in the loop.

\paragraph{Contribution.} We present a suite of five synthetic,
schema-faithful Atlassian environments, the verifiable rewards that score
them, prompted baselines on the same rewards, and end-to-end RL training
results. Concretely:
\begin{itemize}
\item Environments that emulate the Jira REST v3 and Confluence v2
  endpoints used in common automation flows (issue read/create, page
  read/create, transitions, labels), with stateful parent--child
  constraints and per-batch reset (Section~\ref{sec:env}).
\item Verifiable reward functions that decompose into per-argument
  correctness, structural bonuses for the validate--mutate--verify
  pattern, and penalties for hallucinated or duplicate calls --- all
  computable from the tool-call trace (Section~\ref{sec:reward}).
\item Prompted baselines for Qwen3-1.7B and Qwen3.5-4B, run via the
  HuggingFace Inference Router against the identical reward checkers used
  at training time (Section~\ref{sec:results}).
\item An end-to-end RLVR training recipe using GRPO~\cite{shao2024deepseekmath}
  on Qwen3 models~\cite{yang2025qwen3} via TRL~\cite{vonwerra2020trl},
  with a convergence-based early-stopping callback
  (Section~\ref{sec:training}), reaching $\geq 0.95$ average reward on
  every non-degenerate scenario in tens to low hundreds of generation
  batches and producing the largest absolute lift over prompted 4B
  baselines on schema-heavy creation tasks (e.g.\ $+0.65$ on Confluence
  page creation).
\end{itemize}

The result is a proof of concept: the recipe works on this cheap
substrate, and Section~\ref{sec:limits} foregrounds what it does not yet
establish --- chiefly that hand-crafted verifiable rewards do not scale to
the full Atlassian public surface, and that one of our five scenarios has a
saturating reward shape the prompted 4B already maxes out, which we keep as
a transparency control rather than part of the headline claim.

\section{Related Work}
\label{sec:related}

\paragraph{Tool-use agents.} Toolformer~\cite{schick2023toolformer} and
ToolLLM~\cite{qin2024toolllm} train models to invoke real APIs;
ReAct~\cite{yao2023react} and CodeAct~\cite{wang2024codeact} interleave
reasoning with action. These works prove that LLMs can use tools but
largely rely on supervised data; we focus on the RL phase and on
training-time environments that obviate live calls.

\paragraph{RL with verifiable rewards.} DeepSeek-R1~\cite{deepseek2025r1},
DeepSeekMath~\cite{shao2024deepseekmath}, and T\"ulu 3~\cite{lambert2024tulu3}
demonstrate that programmatically checkable rewards drive strong reasoning
gains, primarily on math and code. Concurrent work extends the recipe to
tool-use and agentic settings: ToolRL~\cite{qian2025toolrl} argues that
reward design is the load-bearing piece of tool-use RL, and
Agent-RLVR~\cite{da2025agentrlvr} trains software-engineering agents with
environment rewards and guidance. Our contribution sits in the same family
of methods but is the environment-and-reward design for a different
surface: narrow enterprise SaaS APIs where the verifier inspects nested
argument values and call ordering rather than a final numeric answer, at
sub-second latency and with no live API in the training loop.

\paragraph{Agent benchmarks and synthetic worlds.}
WebArena~\cite{zhou2024webarena}, AgentBench~\cite{liu2024agentbench}, and
SWE-bench~\cite{jimenez2024swebench} evaluate agents on rendered web pages
and real GitHub repositories. AppWorld~\cite{trivedi2024appworld} and
ToolSandbox~\cite{lu2024toolsandbox} provide stateful synthetic worlds
spanning many apps. Our work is narrower in surface but tailored for
training: every reward signal is computed at sub-second latency from the
tool-call trace, and the environment fits in a few hundred lines of
Python, making it suitable for the inner loop of RL.

\section{Environment Design}
\label{sec:env}

\subsection{Design principles}
The environments target four properties:

\paragraph{(P1) Schema fidelity.} Tool signatures and response payloads
mirror the public Jira REST v3 and Confluence v2 contracts (e.g.\
\texttt{POST\allowbreak /rest/\allowbreak api/\allowbreak 3/\allowbreak issue}
expects \texttt{fields.\allowbreak project.\allowbreak key},
\texttt{fields.\allowbreak parent.\allowbreak key}, and
\texttt{fields.\allowbreak issuetype.\allowbreak id="10003"} for a sub-task),
so policies trained here speak the same wire format as their live
counterparts.

\paragraph{(P2) Stateful but resettable.} Each environment exposes a
mutable resource pool (issues, pages) so that \texttt{create\_*} calls
have side effects observable by subsequent \texttt{get\_*} calls; a
\texttt{reset\_synthetic\_data()} hook is invoked before every reward
computation to restore the pristine state, preventing state-pollution
across rollouts.

\paragraph{(P3) Determinism.} No randomness in the environment: identical
tool-call sequences yield identical responses. This makes reward
attribution exact and the substrate usable as a regression fixture.

\paragraph{(P4) Verifiability.} The ground-truth solution for each prompt
is a small dictionary of expected argument values, so the reward function
never needs to invoke an oracle LLM or hit a live service.

\subsection{Scenarios}
\label{sec:scenarios}
We instantiate five scenarios spanning Jira, Confluence, and the
cross-product (Table~\ref{tab:scenarios}). All scenarios share a common
interaction pattern: the user prompt names a resource by key, the agent is
expected to validate that the resource exists, mutate state with a write
call, and optionally verify the result with a final read.

\begin{table*}[t]
\caption{Scenarios in the synthetic Atlassian suite. Tools are surfaced to
the model via the OpenAI-compatible function-calling interface; the
synthetic data layer holds 6--12 prompts per scenario with hand-written
ground-truth arguments.}
\label{tab:scenarios}
\scriptsize
\begin{tabular}{@{}p{2.0cm}p{3.9cm}p{5.0cm}p{4.4cm}@{}}
\toprule
\textbf{Scenario} & \textbf{Tools} & \textbf{State / constraints} & \textbf{Reward signal}\\
\midrule
Jira ticket transition &
\texttt{get\_issue\_summary\_and\_description}, \texttt{get\_transitions}, \texttt{transition\_issue} &
6 issues, each with 3 named transitions and a target final state &
Correct transition id + tool ordering\\
\addlinespace[1pt]
Jira sub-task creation &
\texttt{get\_issue}, \texttt{create\_issue} &
2 parent issues, 3 users; \texttt{create\_issue} validates parent and project &
Per-field correctness on summary, parent, project, assignee, issuetype\\
\addlinespace[1pt]
Confluence page creation &
\texttt{get\_page}, \texttt{create\_page} &
3 root pages across 3 spaces; \texttt{create\_page} validates parent-in-space &
Per-field correctness on title, parentId, spaceId, body\\
\addlinespace[1pt]
Confluence page labeling &
\texttt{get\_page\_content}, \texttt{get\_page\_labels}, \texttt{add\_labels}, \texttt{search\_pages} &
6 pages with realistic technical/HR/ops bodies, hand-labelled gold set &
Recall of expected labels + workflow ordering\\
\addlinespace[1pt]
Cross-product (Jira+Conf.) &
All of \texttt{get\_issue}, \texttt{create\_issue}, \texttt{get\_page}, \texttt{create\_page} &
Merged state pool; a single prompt requires both a sub-task and a page &
Per-platform correctness + cross-platform completion bonus\\
\bottomrule
\end{tabular}
\end{table*}

\subsection{Example tasks}
\label{sec:examples}
One user prompt per scenario, drawn verbatim from the training set:
\begin{itemize}
\item \textit{Ticket transition.} ``Please resolve ticket ISK1.''
\item \textit{Sub-task creation.} ``Create a sub-task `Write QA test
  cases' under ABC-123 and assign to admin123.''
\item \textit{Page creation.} ``Create a page `Oncall Runbook -
  Payments' under parent page 789 in space PAY.''
\item \textit{Page labeling.} ``Can you correctly label page 10001?''
\item \textit{Cross-product.} ``Create a sub-task `Write Release Notes'
  under ABC-123 and assign to admin123. Also create a Confluence page
  `Release Notes 1.0' under parent 789 in space PAY.''
\end{itemize}

At convergence on the cross-product scenario the agent reads the named
parent issue and the named parent page, then issues both create calls
with correct nested arguments; the reward function
(Section~\ref{sec:reward}) attributes credit for each correct argument and
for the read-before-write ordering. The converged tool-call sequence is
reproduced in Appendix~\ref{app:rollouts} (Listing~\ref{lst:l3}).

\section{Verifiable Reward Functions}
\label{sec:reward}

For each scenario we hand-design a dense reward function that maps the list
of tool calls in a rollout to a scalar in $[0, 1]$. The reward decomposes
into three additive groups:

\paragraph{(R1) Per-argument correctness.} For each expected argument in
the gold dictionary, we award a small constant (0.10 in the cross-product
reward; 0.10--0.15 per slot in the single-product creation rewards) if the
agent's emitted argument matches. This is the dominant signal: e.g., in
the sub-task scenario, five fields (\texttt{summary}, \texttt{parent.key},
\texttt{project.key}, \texttt{assignee}, \texttt{issuetype.id}) each
contribute 0.10, for a correctness ceiling of 0.50. For Confluence page
labeling, R1 is instead a per-label recall term: 0.10 for every gold label
the agent eventually adds (Table~\ref{tab:scenarios}).

\paragraph{(R2) Structural bonuses.} We award the validate--mutate--verify
pattern: a \texttt{get\_*} call before any \texttt{create\_*} call, the
correct write next, and an optional \texttt{get\_*} after the create to
confirm. The cross-product reward awards 0.05 per step; the single-product
creation rewards tune these higher (typically 0.05--0.20 per step) so that
an isolated structural signal still drives learning when the per-argument
ceiling is smaller. Bonuses are gated by per-platform ``valid state''
predicates so that they cannot be earned without producing a usable
mutation.

\paragraph{(R3) Penalties.} We subtract for behaviour that would be costly
or destructive against a real API: missing required \texttt{create\_*}
calls, invalid payload shape, duplicate creates, hallucinated tool names,
and excessive call counts beyond a per-scenario budget. Penalty magnitudes
are scenario-tuned (e.g.\ $-0.25$ for a missing create in the
cross-product reward, $-0.30$ in the sub-task reward); the cross-product
values are listed in Table~\ref{tab:reward}. The final reward is clamped
to $[0, 1]$.

Table~\ref{tab:reward} decomposes the cross-product reward, the most
complex; the single-product rewards follow the same R1/R2/R3 template with
per-scenario magnitudes, and ticket-transition is the known outlier
(Section~\ref{sec:limits}).

\begin{table}[t]
\caption{Cross-product reward decomposition. Maximum positive budget is
1.35 before clamping; penalties are subtracted before clamping to
$[0, 1]$.}
\label{tab:reward}
\small
\begin{tabular}{@{}lr@{}}
\toprule
\textbf{Component} & \textbf{Value}\\
\midrule
Jira per-arg correctness ($5 \times 0.10$) & 0.50\\
Confluence per-arg correctness ($4 \times 0.10$) & 0.40\\
Jira structure: get-before-create, verify-after & 0.15\\
Confluence structure: get-before-create, verify-after & 0.15\\
Cross-platform completion bonus (gated on both) & 0.15\\
\midrule
Missing \texttt{create\_issue} / \texttt{create\_page} & $-0.25$ each\\
Invalid Jira / Confluence payload & $-0.25$ each\\
Duplicate \texttt{create\_issue} / \texttt{create\_page} & $-0.15$ each\\
Hallucinated tool name & $-0.15$ each\\
Excess calls beyond budget ($> 6$) & $-0.05$ each\\
\bottomrule
\end{tabular}
\end{table}

\section{Training Setup}
\label{sec:training}

We train Qwen3-1.7B~\cite{yang2025qwen3} (ticket transition) and
Qwen3.5-4B~\cite{qwen2026qwen35} (all other scenarios) with
GRPO~\cite{shao2024deepseekmath} as implemented in TRL~\cite{vonwerra2020trl}.
We use 4--16 generations per prompt (4 for the larger-prompt-budget
scenarios; 16 for ticket-transition and labeling, which expose fewer
distinct prompts per batch), two optimisation iterations per batch, BF16
weights, gradient accumulation of 4, and a maximum completion length of
2048--4096 tokens depending on scenario. A reward-convergence callback
stops training once the epoch-average reward delta falls below 0.01 for a
per-scenario patience window (5--10 epochs), rotating checkpoints to retain
the most recent two on disk.

\paragraph{Hardware.} The 1.7B runs fit on a single Nvidia RTX PRO 6000
Blackwell; the 4B runs use two --- intentionally sized for a small on-prem
or single-node cloud budget.

\section{Results}
\label{sec:results}

\begin{figure*}[t]
\centering
\begin{minipage}[t]{0.49\textwidth}
\centering
\includegraphics[width=\linewidth]{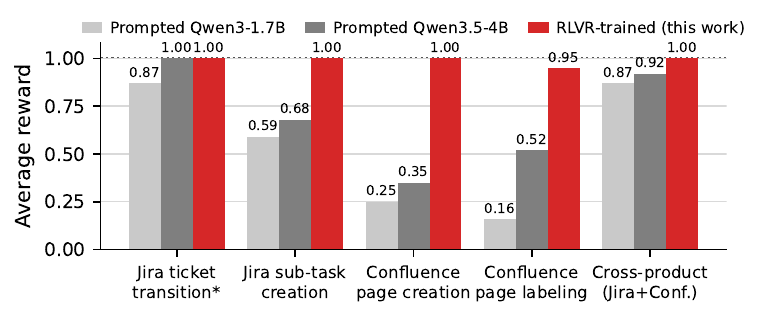}
\caption{Prompted-baseline vs.\ RLVR-trained average reward per scenario,
scored by the identical reward. Baselines (Qwen3-1.7B, Qwen3.5-4B) run via
the HuggingFace Inference Router; ``RLVR-trained'' is the final-epoch GRPO
average. Ticket-transition (asterisked) saturates trivially and is not
informative on its own (Section~\ref{sec:limits}).}
\Description{Grouped bar chart of average reward for five scenarios, with
three bars per scenario: prompted Qwen3-1.7B, prompted Qwen3.5-4B, and the
RLVR-trained policy. The RLVR-trained bars reach 0.95--1.00 across
scenarios, above both baselines except on ticket transition where the 4B
baseline already saturates.}
\label{fig:bars}
\end{minipage}
\hfill
\begin{minipage}[t]{0.49\textwidth}
\centering
\includegraphics[width=\linewidth]{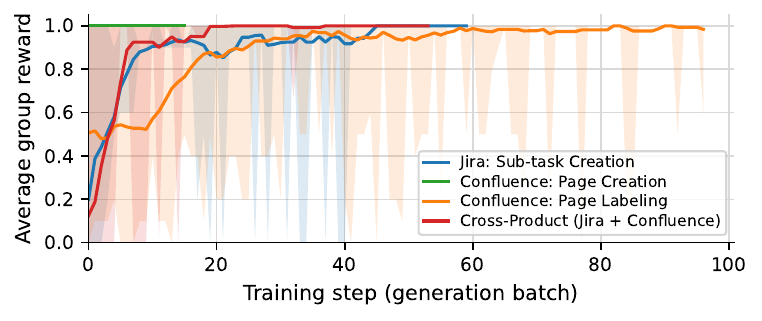}
\caption{Average group reward vs.\ generation-batch index, four scenarios
trained with GRPO on Qwen3.5-4B. Solid lines: 5-batch moving average;
shaded bands: per-batch min/max over the four-generation group. All four
cross 0.95 within $\leq 100$ batches.}
\Description{Line plot of average group reward versus training step for
four scenarios. Cross-product and sub-task creation rise fastest, page
labeling rises more gradually to about 0.95, and page creation stays flat
near 1.0; shaded min/max bands narrow as training proceeds.}
\label{fig:dynamics}
\end{minipage}
\end{figure*}

\subsection{Prompted baselines vs.\ RLVR-trained}
We score prompted Qwen3-1.7B and Qwen3.5-4B against the same reward
functions used during training, run via the HuggingFace Inference Router
(temperature 0.7, 25-turn cap), with synthetic state reset between every
task and tool-call arguments normalised to dicts so baseline scores are
directly comparable. Figure~\ref{fig:bars} reports the average reward per
scenario. We keep ticket-transition only as a transparency control --- its
reward saturates (Section~\ref{sec:limits}), so the prompted 4B already
scores 1.00; the four non-degenerate scenarios carry every claim below.

On the four scenarios whose rewards are non-degenerate, the RL-trained
policy beats prompted Qwen3.5-4B by a meaningful margin. The largest
absolute lifts come on the most schema-heavy creation tasks: Confluence
page creation jumps from 0.35 to 1.00 ($+0.65$), Confluence labeling from
0.52 to 0.95 ($+0.43$), and Jira sub-task creation from 0.68 to 1.00
($+0.32$). On the cross-product task, where the 4B baseline is already
strong at 0.92, RL closes the residual tail to 1.00 --- every one of the
12 evaluation prompts becomes a perfect rollout rather than 8 of 12.
Failure modes in the prompted baselines mirror the objective mismatch
flagged in Section~\ref{sec:intro}: plausible-looking calls with the wrong
nested key, dropped required fields, and early-stopping after the read.

\subsection{Training dynamics}
Figure~\ref{fig:dynamics} plots the GRPO reward trajectories
(ticket-transition is omitted; its logging schema was coarser). The three
from-scratch scenarios reach $\geq 0.95$ in under 100 generation batches;
the Confluence page-creation curve is a warm-started stability trace (flat
near 1.0 across 16 batches). The shaded bands report per-batch min/max
across the four-generation group, and contract over training as the policy
stops emitting low-reward rollouts.

Appendix~\ref{app:rollouts} contrasts the two dominant pre-training
failure modes (early stop after the read calls; mis-shaped nested parent
arg followed by a chain of duplicate \texttt{create\_*} retries) against
the converged trace. These traces are human-inspectable evidence that the
gain is behavioural, not reward-hacking: the converged policy reads before
it writes, shapes nested arguments correctly, and stops without
hallucinated or duplicate calls --- visible in the call sequence,
independent of the scalar.

\section{Limitations and Future Work}
\label{sec:limits}

\boldparagraph{Narrow prompt coverage; no held-out evaluation.} Each scenario
ships with 6--12 training prompts. Convergence to $\geq 0.95$ average
reward on those prompts shows that GRPO with the proposed verifiable
reward can fit the training distribution; it does not measure
generalisation to unseen prompts. A held-out set --- same schema, new
entities, summaries, and parent--child combinations --- is the immediate
next step.

\boldparagraph{Reward-shape artifacts.} The ticket-transition reward awards
$+0.25$ per tool call, clamped at 1.0, with no penalty for excess or
hallucinated calls, so the prompted 4B already saturates at 1.00 (one
rollout invoked \texttt{transition\_issue} nine times and still maxed
out). A separate type bug compares the gold transition id as an integer
against a string-typed schema, denying the schema-compliant 1.7B a bonus
the 4B collects. We keep these shapes verbatim --- they are the rewards
under which the rollouts were generated --- but flag per-scenario
reward-hacking audits as unfinished work.

\boldparagraph{Reward engineering does not scale by hand.} Each reward was
hand-tuned through pilot GRPO loops that exposed one of three failure
modes: no convergence (signal too weak), reward hacking, or premature
plateau (group-relative advantages collapse on a partial solution).
Scaling this to hundreds of Atlassian endpoints is the load-bearing
scalability question. Much of each reward is mechanically derivable,
though --- R1 per-argument terms from a schema's \texttt{required} fields,
R3 type penalties from its \texttt{type}/\texttt{enum} constraints, leaving
only R2 structural bonuses to hand-tune --- so OpenAPI-driven reward
synthesis, checked by pilot-run signal or an LLM-as-reward-designer, is the
most important follow-up.

\boldparagraph{Live-API transfer.} Schema fidelity at the wire format does not
guarantee robustness to real-world failure modes (rate limits, eventual
consistency, permission errors, partial responses); the synthetic
environments are intended as a cheap inner loop paired with a smaller
live-tenant validation loop: record--replay of captured responses, fault
injection (429/403, stale eventual-consistency reads), and a low-volume
sandbox-tenant canary.

\boldparagraph{Stronger baselines and open questions.} Our prompted
baselines hold the base model fixed, isolating the post-training objective;
they are a controlled test, not a claim that RL is the only route. SFT on
successful traces, ReAct- or plan-execute-style scaffolding, and a frontier
model with a Python REPL over these environments are the next rungs ---
separating whether RL is necessary from merely sufficient --- alongside an
order-of-magnitude more prompts, larger base models, and multi-turn user
simulators.

\paragraph{Reproducibility.} Rewards are deterministic functions of the
tool-call list, so any rollout re-scores post-hoc against the same checker;
Tables~\ref{tab:scenarios}--\ref{tab:reward}, Sections~\ref{sec:examples}
and~\ref{sec:training}, and the Appendix~\ref{app:rollouts} rollouts specify
the substrate at the level needed to reimplement it.

\bibliographystyle{ACM-Reference-Format}
\bibliography{references}

\appendix

\section{Pre- and Post-Training Tool-Call Sequences}
\label{app:rollouts}

The listings below are the ordered tool-call sequences (tool name and
argument dictionary) emitted by the policy during a single cross-product
rollout, lifted from \texttt{atlassian\_rollouts.json} with tool responses
and intermediate reasoning omitted. Listings~\ref{lst:l1} and~\ref{lst:l2}
are representative pre-convergence failure modes; Listing~\ref{lst:l3} is a
converged success.

\paragraph{Failure mode A: early stop.} Listing~\ref{lst:l1} is a
generation-batch 0 rollout (reward $= 0.00$). The policy correctly reads
both parent resources, then halts before issuing the two required write
calls. This is the dominant pre-training failure mode in the cross-product
scenario and the one captured by the qualitative claim in
Section~\ref{sec:results}.

\begin{lstlisting}[style=toolcall,label={lst:l1},caption={Pre-training rollout, step 0, reward = 0.00. The policy stops after the read calls and never emits a \texttt{create\_*}.}]
# user: Create 'Deploy Staging' sub-task under XYZ-456
# for admin123. And a 'Staging Environment' page
# under 100 in ENG.
get_issue(issueIdOrKey="XYZ-456")
get_page(page_id="100")
# (policy ends turn)
\end{lstlisting}

\paragraph{Failure mode B: structural OK, nested args wrong.}
Listing~\ref{lst:l2} is a generation-batch 2 rollout (reward $= 0.52$).
The policy emits the full read-then-write sequence, but the first
\texttt{create\_issue} passes \texttt{parent} as a bare string
\texttt{"ABC-123"} instead of \texttt{\{"key": "ABC-123"\}}, the Jira-v3
wire format. The policy then issues five more \texttt{create\_issue} calls
trying alternate shapes (string parent, dropped assignee,
\texttt{projectKey} instead of \texttt{project.key}); each collects the
duplicate-create penalty.

\begin{lstlisting}[style=toolcall,label={lst:l2},caption={Pre-training rollout, step 2, reward = 0.52. Five duplicate \texttt{create\_issue} calls probe alternate argument shapes after the first fails the payload-shape check.}]
# user: Create a sub-task 'Write Release Notes' under
# ABC-123, assignee admin123. Also create a
# Confluence page 'Release Notes 1.0' under
# parent 789 in space PAY.
get_issue(issueIdOrKey="ABC-123")
get_page(page_id="789")
create_issue(fields={"project": {"key": "ABC"},
    "parent": "ABC-123",  # wrong shape
    "summary": "Write Release Notes",
    "issuetype": {"id": "10003"},
    "assignee": {"accountId": "admin123"}})
create_page(spaceId="PAY", title="Release Notes 1.0",
    parentId="789",
    body={"representation": "storage",
        "value": "<h2>Release Notes 1.0</h2>..."})
create_issue(fields={..., "parent": "ABC-123",
    "assignee": "admin123"})  # wrong
create_issue(fields={..., "parent": "ABC-123"})  # wrong
create_issue(fields={"projectKey": "ABC",
    ..., "parent": "ABC-123"})  # wrong
create_issue(fields={..., "parent": "ABC-123"})  # wrong
\end{lstlisting}

\paragraph{Success: converged.} Listing~\ref{lst:l3} is a
generation-batch 53 rollout (reward $= 1.00$). The policy reads both
parents, then issues exactly one \texttt{create\_issue} and one
\texttt{create\_page} with correct nested shape --- including the
\texttt{issuetype.id="10003"} subtask discriminator that the system prompt
names but does not demonstrate.

\begin{lstlisting}[style=toolcall,label={lst:l3},caption={Post-training rollout, step 53, reward = 1.00. Read both parents, then one correctly shaped write per platform.}]
# user: Create a sub-task 'Write Release Notes' under
# ABC-123, assignee admin123. Also create a
# Confluence page 'Release Notes 1.0' under
# parent 789 in space PAY.
get_issue(issueIdOrKey="ABC-123")
get_page(page_id="789")
create_issue(fields={
    "project": {"key": "ABC"},
    "parent": {"key": "ABC-123"},
    "summary": "Write Release Notes",
    "issuetype": {"id": "10003"},
    "assignee": {"accountId": "admin123"}})
create_page(spaceId="PAY", title="Release Notes 1.0",
    parentId="789",
    body={"representation": "storage",
        "value": "<p>...</p>"})
\end{lstlisting}

\end{document}